\documentclass[conference]{IEEEtran}
\usepackage{amsmath}
\usepackage[]{algorithm2e}
\usepackage{graphicx}
\usepackage{amssymb}
\usepackage{algorithmic}
\usepackage{float}
\usepackage{caption}
\usepackage{subcaption}
\usepackage{bbm}

\ifCLASSINFOpdf
\else
\fi
\begin{document}

\title{Similarity Search Over Graphs Using Localized Spectral Analysis}

\author{\IEEEauthorblockN{Yariv Aizenbud}
    \IEEEauthorblockA{School of Mathematics \\
        Tel Aviv University, Israel\\
    }
\and
\IEEEauthorblockN{Amir Averbuch}
\IEEEauthorblockA{School of Computer Science\\
Tel Aviv University, Israel\\
}
\and
\IEEEauthorblockN{Gil Shabat}
\IEEEauthorblockA{School of Computer Science\\
    Tel Aviv University, Israel\\
}
\and
\IEEEauthorblockN{Guy Ziv}
\IEEEauthorblockA{School of Mathematics\\
    Tel-Aviv University, Israel\\
}}

\maketitle

\begin{abstract}
This paper provides a new similarity detection algorithm. Given an
input set of multi-dimensional data points\footnote{Each data point
is assumed to be multi-dimensional} and an additional reference data
point for similarity finding, the algorithm uses kernel method that
embeds the data points into a low dimensional manifold. Unlike other
kernel methods, which consider the entire data for the embedding,
our method selects a specific set of kernel eigenvectors. The
eigenvectors are chosen to separate between the data points and the
reference data point so that similar data points can be easily
identified as being distinct from most of the members in the
dataset.
\end{abstract}

\IEEEpeerreviewmaketitle

\section{Introduction}
In recent years, there is an on-going interest in finding efficient
solutions to discover similarity between data points. Measuring
similarity plays a central role in computer
vision~\cite{wang2014hashing}, speech recognition, text
analysis~\cite{broder1997resemblance} and anomaly
detection~\cite{chandola2009anomaly}, to name some. The problem can
be defined as follow: Given a reference data point from a collection
of data points in an $n$-dimensional metric space, we want to find
the most similar data points with respect to the reference data
point. While the interest in solving the similarity search problem
accurately has always been an important goal, the rapid growth in
the amount of collected data raises a need for solving this problem
efficiently as well. In this work, we propose a robust method for
similarity detection using the notion of localized spectral methods
over graphs~\cite{cheng2016diffusion}. The main advantage of our
method lies in the fact that while many algorithms, such as
nearest-neighbors and its variants, search for similarity in the
feature space (whether in the original ambient space or in the
low-dimensional embedded space), our method searches for similarity
in the intrinsic characteristics. This is done by looking at the
eigenvectors that enable us to  separate between the relevant data
points and the rest of the data. This methodology consists of two
steps: 1. Decomposition of a graph (given as a kernel) that
represents the imposed similarity metric between data points. 2. A
search for resemblance in the relevant new space where separation
exists following 1. We apply this method to synthetic and real
datasets and compare the obtained results with other known methods.

\section{Preliminaries}
\subsection{Related Work}
Similarity search has a key role in many applications involving
high-dimensional data. Extensive research is done to achieve both
efficient and accurate similarity search results. Nearest-neighbors
search and its approximated variants such as Hashing methods
~\cite{gionis1999similarity}, are a popular solutions and have been
widely used to achieve fast approximate similarity search. 
Others like~\cite{song2014robust,kraus2016nearbucket} implement more
robust and efficient methodologies. The similarity search task
in~\cite{shen2007image} is done  by redefining the feature space via
local intensity histograms. This can be further used as attributes
for image matching. Another way is to construct the eigenvectors for
each pixel by geometrical moments based on local histogram
 to automatically detect the corresponding landmark in CT brain images~\cite{sun2009eigenvector}.
 Most methods address the similarity search problem in feature space by using either the original feature space or by an alternative representation
 including hashing and dimensionality reduction techniques. In this work,
  we address this problem by using the intrinsic characteristics of the data and not the feature space as commonly used.

\subsection{Geometry Preservation by Kernels}

Data analysis often involves non-linear relations between data
points that are harder to extract via conventional linear methods.
PCA and SVM, for example, are two well known methods that lack the
ability to handle such relations between data points due to their
linear nature. As a direct result, one would like to choose a method
that allows the data points to be mapped into a higher dimensional
space while exploiting the non-linear properties and relations.
Kernel methods enable us to operate and analyze data in a
high-dimensional environment while extracting the non-linear
properties and relations in different scenarios
\cite{bishop2006pattern}. When data is analyzed to find similarities
between data points, exploiting non-linearity is important and
therefore kernel methods can be useful. Two important examples in
the area of kernel methods are Diffusion Maps~\cite{coifman2006a}
and Laplacian Eigenmaps~\cite{belkin2003laplacian}. Diffusion Maps
show that the eigenvectors of Markovian matrix can be considered as
a set of coordinates of the dataset, which can be represented as a
set of data points in a Euclidean space.  This procedure captures
most of the original geometry of the data. Laplacian Eigenmaps show
that a neighborhood-information-based graph can be considered as a
discrete approximation of the low-dimensional manifold in the
high-dimensional space. The usefulness of kernel methods and its
relation to dimensionality reduction, classification and anomaly
detection are described in
~\cite{mishne2013multiscale,chernogorov2011detection,du2012discriminative}.

\section{Main Approach}
\subsection{Similarity Search Assumptions}
Our method relies on two main assumptions. 1. There is a
low-dimensional space, which separates between data points,
specifically it separates between our reference data point and the
rest of the data points. 2. Each data point, which belongs to a
high-dimensional space, can be characterized in a lower dimensional
space than the original (ambient) space by choosing an appropriate
kernel. Our first assumption, which maps the data into a
low-dimensional space, is common since in most cases there is a
strong dependency (linear or non-linear) between different
coordinates. This results in a lower dimensional space than the
ambient space. If the data points are inseparable then the data is
assumed to be homogeneous. Hence, the notion of similarity is
meaningless. Choice of an appropriate kernel in our second
assumption can uncover hidden relations between data points.  Our
approach does not rely on prior knowledge or assumptions regarding
the data/paramters distribution.

\subsection{Similarity Search Description}
Data points (both via linear or non-linear methods) are mapped into
their low-dimensional embedding space by utilizing  the largest
eigenvalues and their corresponding eigenvectors. This process
captures the geometry of the data. Using successfully  a  small
number of eigenvalues is demonstrated in \cite{coifman2006a}. 
 Classical spectral
methods suggest to use the largest eigenvalues and their
corresponding eigenvectors. In contrast to commonly use of
eigenvalues, according to the above methods, we will not
characterize accurately  the reference point and its similar data
points.

We suggest a method that is classified as a localized spectral
methods over graphs. The method proposes that the intrinsic
characteristics of a reference data point can be measured mostly by
its top eigenvectors values. 
We define the top eigenvectors to have the largest absolute value in the coordinate of the reference data
point. 
Moreover, data points which have similar top eigenvectors
 have shared characteristics. Similarity is defined by the
norm of the localized spectral reconstruction error. We test this
method on both synthetic and real datasets and provide comparable
results.

\subsection{Localized Spectral Similarity Analysis Algorithm}
Let $X = \{x_1, \ldots, x_m\}$ be a set of data points in $R^n$ and
let $x_r$ be a reference data point. We are looking to identify data
points from $X$  that are similar to $x_r$. We build a graph $G =
(X, K)$ where $K = k(x_i,x_j)$. The normalized kernel is $P = D^{-1}K$ where $D$ a diagonal matrix, with $D_{ii} = \sum_{j}K_{ij}$.
Each row of $P$ is summed to 1. It consists of only real entries,
therefore the matrix can be viewed as a Markov transition matrix. We
define A to be the matrix $A = D^{-\frac{1}{2}}KD^{-\frac{1}{2}}$, which is a symmetric matrix that has $m$ positive eigenvalues that can be viewed as a graph Laplacian matrix. The eigenvalue decomposition (EVD) of $A$ is donated by $U \Sigma U^T$. $U_r=U(r,:)$ is the $r$th
row of the matrix $U$, which is the coordinates of $x_r$ on the embedded axes. $\tilde{U}_r$ is the absolute values of the vector $U_r$ sorted
in descending order. Denote
$T=\Big[\tilde{U}_1,\ldots,\tilde{U}_m \Big]$. The matrix $T$
contains the eigenvectors ($U$) sorted by descending significance to
the reference data point $x_r$ and let $T_k$ be the truncated matrix
that consists of the first $k$ columns of $T$, $T_k=T(:,1:k)$.

Next, we enhance data points similar to the reference data point
$x_r$ by correlating the embedded data points $T_k$ with a
unit-vector in the direction of $U_r$, i.e.
\begin{equation}
S = T_k \frac{\tilde{U}^T_r(1:k)}{\Vert \tilde{U}_r(1:k) \Vert}
\end{equation}
Once $S$ is computed, each data point $x_i$ gets a score $\vert
s_i\vert$, where similar data points has a higher score $\vert s_i
\vert$.

\begin{algorithm}[!ht]
    \caption{Find Kernel Similarities}
    \label{alg:KernelSim}
    \textbf{Input:}  Data matrix $X\in \mathbb{R}^{m\times n}$ with $m$ measurements and $n$ features, $k$ - number of vectors to use, $r$ - point index to search for\\
    \protect
    \textbf{Output:} Similarity score for each data point
    \begin{algorithmic}[1]
        \STATE Build kernel matrix $K$
        \STATE Construct Diffusion map $A = D^{-\frac{1}{2}}KD^{-\frac{1}{2}}$, $D_{ii}=\sum_j K_{ij}$.
        \STATE Compute the EVD of $A$, $A=U\Sigma U^T$
        \STATE Sort by descending order $\vert u_{rj} \vert$, $j=1,\ldots,m$ and store the $k$ indexes (indicated by $j_1,\ldots,j_k$) that correspond to the largest values of $\vert u_{rj}\vert$. $\tilde{U}_r=\Big[\vert u_{rj_1}\vert,\ldots,\vert u_{rj_m}\vert \Big]$
        \STATE Form a new data matrix $T$ on the new vectors: $T \leftarrow \left[\tilde{U}_1 \ldots \tilde{U}_m\right]$, $T_k \leftarrow T\left(:,1:k\right)$
        \STATE Compute the score vector, $S=T_k\frac{\tilde{U}^T_r(1:k)}{\Vert \tilde{U}_r(1:k)\Vert}$
        \RETURN the absolute value of the elements in  $S$
    \end{algorithmic}\
\end{algorithm}
In practice, since $m$ is large, it is usually impractical to compute the full EVD of the kernel $K$. Therefore, one can compute only the first largest $l$ eigenvectors for $k\ll\l \ll m$. This can be done, for example, by using power iterations \cite{golub} or by randomized SVD algorithms \cite{aizenbud2016matrix, halko2011finding, WLRT}.

\section{Experimental Results}
Algorithm \ref{alg:KernelSim} was tested on two datasets:
1) A synthetically generated 3D surface and 2) An image of size $256 \times
256$.\newline
The 3D surface in the first experiment was injected by  two abnormal
data points that are hovered above the surface. One point was
selected to be the reference data point while the other as test data
point. The selection of the two data points was random. In the
second experiment - the Mona Lisa's painting, the image was
rearranged into a series of columns from sliding blocks of size
$3\times 3$ and a selected reference patch of size $3\times3$ from
the Mona Lisa's skin was randomly chosen. The aim of both
experiments is to locate the resembling data points, meaning for the
3D figure we would like to find second abnormal data points and for
the Mona Lisa's photo we would like to recognize other patches of
skin.

\begin{figure}[H]
	\centering\text{Generated 3D Surface Experiment}\par\medskip
	\centering
	\begin{subfigure}[t]{0.23\textwidth}
		\includegraphics[width=4.2cm, height=4.2cm]{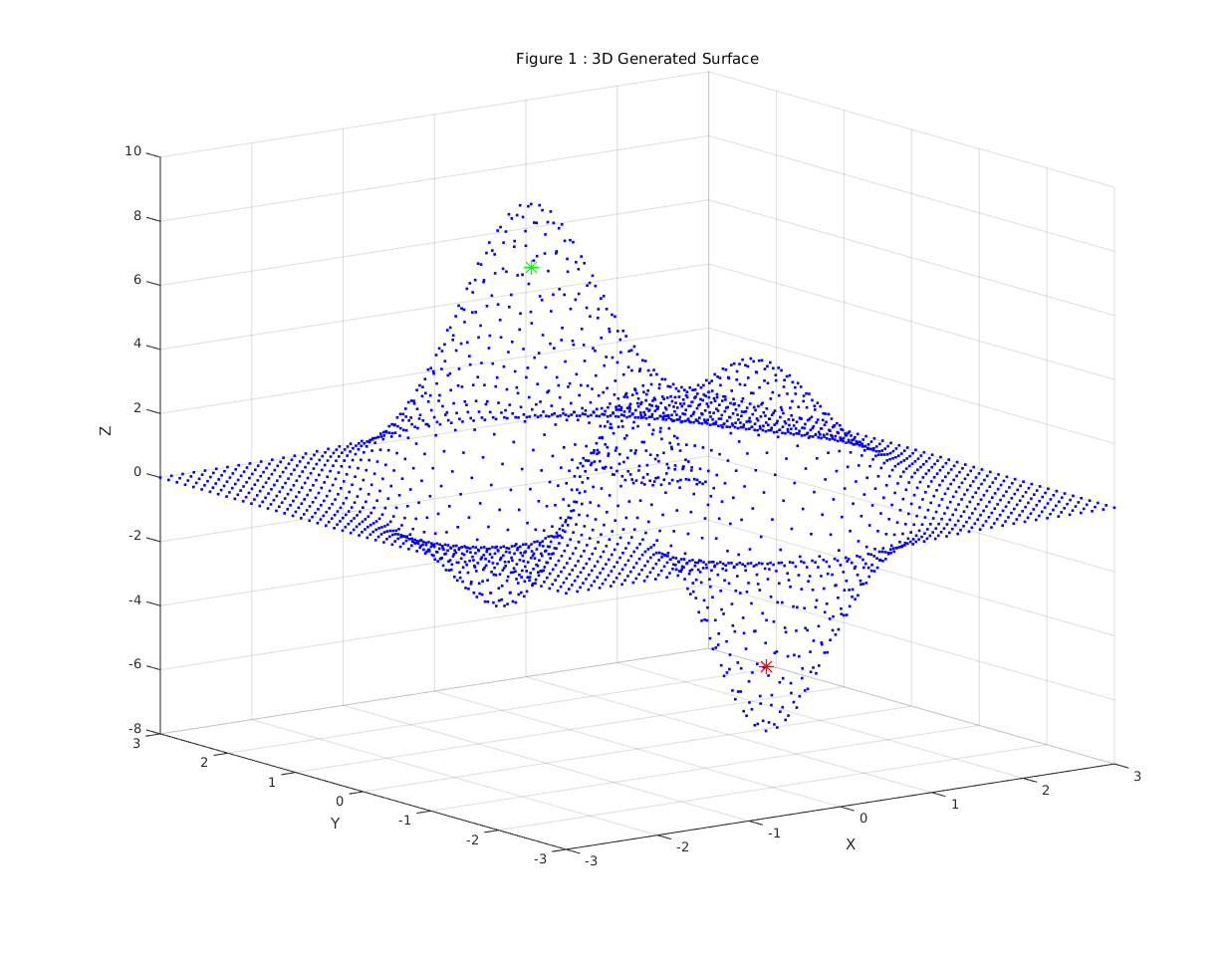}
		\caption{3D surface generated data with two abnormal data points marked in green and red as the reference data point and the desired matching point accordingly. The proposed method find the relevant data point quite easily while alternative methods suggest incorrect data points as similar ones.}
		\label{fig:heals_left}
	\end{subfigure}
	~
	\begin{subfigure}[t]{0.23\textwidth}
		\includegraphics[width=4.2cm, height=4.2cm]{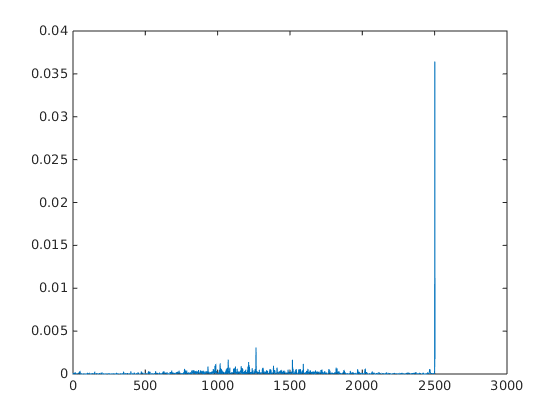}
		\caption{Reconstruction magnitude for each of the data points by using the top three singular vectors of the reference data point with the new corresponding singular values. It is easily seen that the two last data points in the dataset located at rows 2501 and 2502, known as the reference data point and the matching point, both have distinctively high Reconstruction compared to the rest of the data points.}
		\label{fig:heals_right}
	\end{subfigure}
	\caption{ }
	\label{fig:heals}
	
\end{figure}

\subsection{Generated 3D Surface}
Data points which form a 3D surface were generated from 2500 data
points and two new observations outside the terrain were injected to
the data apart from each. Both of the data points were located at
the end of the dataset, in row numbers 2501 and 2502. The goal was to
match the most similar observation to the reference data point. For
comparison purposes Nearest Neighbor (NN) and Kernel Nearest
Neighbor (Kernel-NN) algorithms were chosen~\cite{yu2002kernel}. A Gaussian kernel was selected for this experiment both for the suggested algorithm and for Kernel-NN.
Figure \ref{fig:heals_left} shows the terrain that was generated.
The original data points which belong to the surface are colored in
blue while the two abnormal data points which were artificially
injected into the data are marked in green and red. One of the data
points (green) was chosen to be the reference data point while the
other (red) was chosen to be the matching data point that the
algorithms try to locate. The reference
data point was chosen randomly and we repeated the experiment with
different data point locations over the surface. Figure
\ref{fig:heals_right} shows the reconstruction magnitude for each of
the data points by using the top three singular vectors of the
reference data point with the new corresponding singular values. Note that when taking all $m$ singular vectors instead of choosing the top k ones (top three in the current experiment) the algorithm will not perform well. choosing all $n$ values will construct the space with respect to the full data and not to the reference point and other similar data points which will result in poor performance for similarity detection. Our method outperforms Both NN and Kernel-NN. The alternative
methods had difficulty to find the matching data point in a consistent matter compare to our method for different locations of reference and matching data along the terrain. For the presented experiment (Figure \ref{fig:heals}) Kerenel-KNN ranks the matching data point as 409 most similar while regular KNN only as 2459 out of 2502 data points. Our method ranks it correctly as the most similar data point.

\begin{figure}[H]
	\centering\text{Image Analysis Experiment}\par\medskip
	\centering
	\begin{subfigure}[t]{0.23\textwidth}
		\includegraphics[width=4cm, height=4cm]{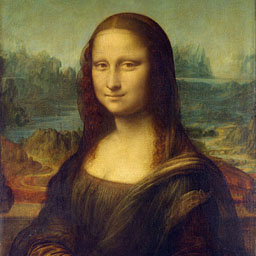}
		\caption{Original Mona Lisa}
		\label{fig:monalisaorig}
	\end{subfigure}
	\begin{subfigure}[t]{0.23\textwidth}
		\includegraphics[width=4cm, height=4cm]{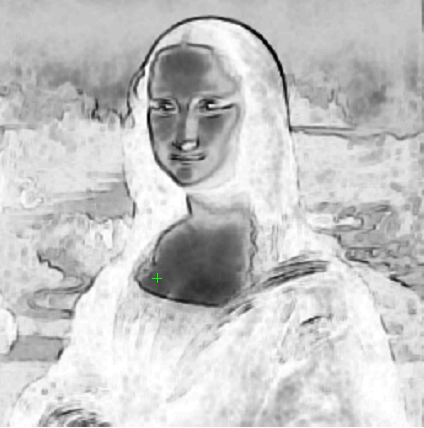}
		\caption{First top eigenvalue of the kernel}
		\label{fig:monalisa1ev}
	\end{subfigure}
	\caption{Dark regions correspond to patches that are similar to the reference patch (indicated by a green $+$) of skin and neck.}
	\label{fig:monalisa}
\end{figure}

\subsection{Image Analysis}

A ($256 \times 256$) pixels Mona-Lisa gray level image was divided into sliding blocks (overlapping elements) of size $3 \times 3$ and later transformed into a data matrix of size $64,516\times 9$, where each row corresponds to an image patch. Next, an arbitrary patch from the Mona Lisa's skin patches was chosen (image coordinate $(166,96)$ indicated by a green $+$ in Figure \ref{fig:monalisaorig}).  Algorithm \ref{alg:KernelSim} was applied to the data matrix, building a Gaussian kernel of size $64,516\times 64,516$ computing the first $15$ eigenvectors. Figure \ref{fig:monalisa1ev} shows the magnitude of the first top eigenvector. It can be seen from the figure, that the dark regions correspond to patches that are similar to the reference patch of skin and neck.

\section{Conclusion}
In this paper, we presented a new algorithm for detecting similarities within a given dataset. The algorithm is based on localized spectral analysis. The method characterizes a reference data point by looking at the significant eigenvectors of the embedding kernel. The significant eigenvectors  form a basis that enables us to differentiate between similar data points from the rest of the data. Numerical results of the algorithms were presented. They exhibit  the potential of the new method.

\bibliographystyle{IEEEtran}
\bibliography{KernelSimilarities}

\end{document}